\documentclass[11pt]{article}
\usepackage[preprint]{acl}

\usepackage{times}
\usepackage{latexsym}
\usepackage[T1]{fontenc}
\usepackage[utf8]{inputenc}
\usepackage{microtype}
\usepackage{inconsolata}
\usepackage{graphicx}
\usepackage{booktabs}
\usepackage{multirow}
\usepackage{arydshln}
\usepackage{pifont}
\newcommand{\cmark}{\ding{51}}
\usepackage{adjustbox}
\usepackage{amsmath}
\usepackage{amssymb}
\usepackage{bm}
\definecolor{c1}{HTML}{003371}
\usepackage{natbib}
\usepackage{caption}
\usepackage{cleveref}
\usepackage[most]{tcolorbox}
\definecolor{blue1}{HTML}{013371}
\newtcbtheorem[]{trainprompt}{Case}%
{colback=gray!00,colframe=blue1,fonttitle=\bfseries, left=.02in, right=.02in,bottom=.02in, top=.02in}{trainprompt}
\usepackage{algorithm}
\usepackage{algorithmic}

\usepackage{newfloat}
\usepackage{listings}
\DeclareCaptionStyle{ruled}{labelfont=normalfont,labelsep=colon,strut=off}
\floatstyle{ruled}
\newfloat{listing}{tb}{lst}{}
\floatname{listing}{Listing}

\title{MathMixup: Boosting LLM Mathematical Reasoning with Difficulty-Controllable Data Synthesis and Curriculum Learning}

\author{
 \textbf{Xuchen Li\textsuperscript{\rm 1,}\textsuperscript{\rm 2,}\textsuperscript{\rm 3,}\textsuperscript{\rm 5}\thanks{Equal contribution.}\thanks{Work done during Xuchen’s internship at ByteDance.}},
 \textbf{Jing Chen\textsuperscript{\rm 5}\footnotemark[1]\thanks{Project leader.}},
 \textbf{Xuzhao Li},
 \textbf{Hao Liang},
\\
 \textbf{Xiaohuan Zhou\textsuperscript{\rm 5}\thanks{Correspondence authors.}},
 \textbf{Taifeng Wang\textsuperscript{\rm 5}},
 \textbf{Wentao Zhang\textsuperscript{\rm 4}\footnotemark[4]}
\\
\textsuperscript{\rm 1}CASIA,
\textsuperscript{\rm 2}ZGCA,
\textsuperscript{\rm 3}UCAS,
\textsuperscript{\rm 4}PKU,
\textsuperscript{\rm 5}ByteDance
\\
 \small{
    s-lxc24@bjzgca.edu.cn, chenjing.celeste@bytedance.com, zhouxiaohuan@bytedance.com, wentao.zhang@pku.edu.cn
 }
}

\begin{document}
\maketitle
\begin{abstract}
In mathematical reasoning tasks, the advancement of Large Language Models (LLMs) relies heavily on high-quality training data with clearly defined and well-graded difficulty levels. However, existing data synthesis methods often suffer from limited diversity and lack precise control over problem difficulty, making them insufficient for supporting efficient training paradigms such as curriculum learning. To address these challenges, we propose MathMixup, a novel data synthesis paradigm that systematically generates high-quality, difficulty-controllable mathematical reasoning problems through hybrid and decomposed strategies. Automated self-checking and manual screening are incorporated to ensure semantic clarity and a well-structured difficulty gradient in the synthesized data. Building on this, we construct the MathMixupQA dataset and design a curriculum learning strategy that leverages these graded problems, supporting flexible integration with other datasets. Experimental results show that MathMixup and its curriculum learning strategy significantly enhance the mathematical reasoning performance of LLMs. Fine-tuned Qwen2.5-7B achieves an average score of 52.6\% across seven mathematical benchmarks, surpassing previous state-of-the-art methods. These results fully validate the effectiveness and broad applicability of MathMixup in improving the mathematical reasoning abilities of LLMs and advancing data-centric curriculum learning.
\end{abstract}

\section{Introduction}
Recent advances in Large Language Models (LLMs), such as DeepSeek R1 \cite{deepseekr1}, Kimi K1.5 \cite{kimik1_5} and OpenAI o3 \cite{gpt4o}, have led to remarkable progress in mathematical reasoning tasks. These models have demonstrated impressive performance across a variety of mathematical benchmarks \cite{tang2024mathscale,math,gsm8k,saxton2019analysing,cao2025large}, narrowing the gap between artificial and human-level problem solving. However, despite these advancements, a critical bottleneck remains: further improvement in LLM mathematical reasoning is fundamentally constrained by the lack of high-quality training data with explicit and controllable difficulty gradients \cite{lu2024mathgenie,li2025verifybench,li2025sciagent}.

\begin{figure}[t!]
  \centering   
  \includegraphics[width=\linewidth]{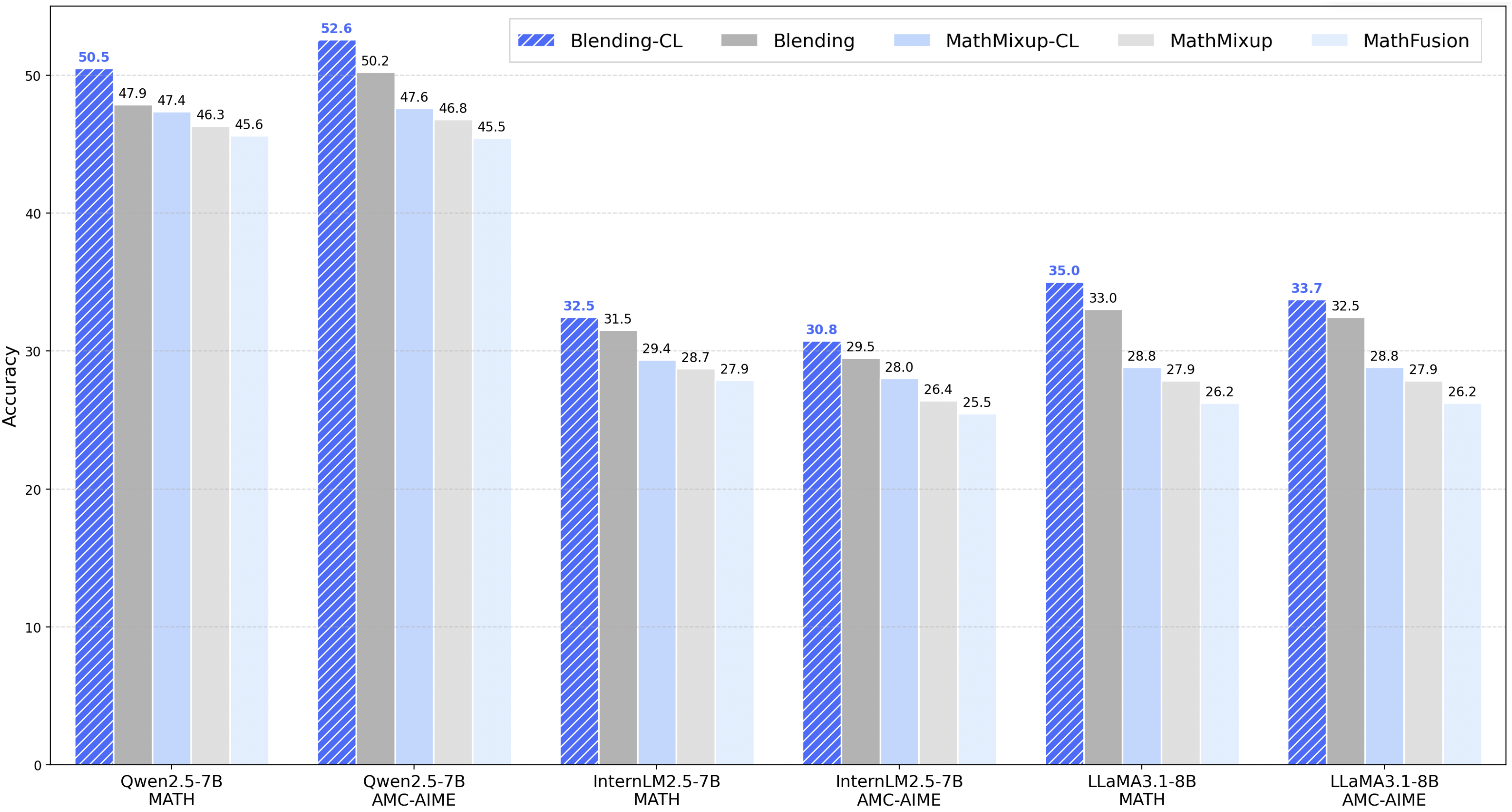}
  \vspace{-20pt}
  \caption{Average accuracy comparison of different datasets and curriculum learning (CL) strategies on seven mathematical reasoning benchmarks across three LLMs (Qwen2.5-7B, InternLM2.5-7B, and LLaMA3.1-8B), with all datasets synthesized separately from MATH and AMC-AIME seeds. MathMixup consistently outperforms the MathFusion baseline, and curriculum learning (MathMixup-CL) further improves performance. Blending MathMixupQA and MathFusionQA (Blending) yields additional gains, while Blending-CL achieves the highest accuracy across all models and settings. These results demonstrate that difficulty-controllable data and curriculum learning are both effective individually, and their combination leads to the greatest improvements in LLM mathematical reasoning.
  }
  \label{fig:introduction}
  \vspace{-20pt}
\end{figure}

Against this backdrop, it is crucial to systematically examine the roles of data quality, diversity, and organization in mathematical reasoning tasks \cite{shen2025let}. In this context, data augmentation and synthesis \cite{wang2024survey,zhou2024jiuzhang3,fedoseev24constraint,chen2025advancing} have become promising approaches to providing high-quality and diverse data. Existing data synthesis methods can be mainly categorized into two types: the first type enhances or rewrites individual problems \cite{metamath,mathscale,mmiqc,refaug}, which, despite increasing the dataset size, often results in generated questions that are highly similar in semantics and difficulty, thus lacking diversity; the second type, exemplified by MathFusion \cite{mathfusion}, merges two similar problems to increase computational complexity. While these methods can increase problem complexity to some extent, they still lack systematic and explicit control over reasoning difficulty, which is essential for robust capability improvement. Meanwhile, curriculum learning \cite{liu2024let,wang2025dump,li2025causalstep} has proven to be an effective paradigm for further improving model reasoning abilities, and its success is fundamentally dependent on the availability of well-graded training data with clear and controllable difficulty progression \cite{chen2025self,zhang2025learning,hu2024fiova}. However, current datasets and synthesis methods are inadequate in providing such difficulty stratification and gradients, which limits the potential of curriculum learning. Therefore, there is an urgent need for a new method capable of generating mathematical reasoning data with controllable and well-graded difficulty, which can not only enhance model capability directly but also facilitate further improvement through curriculum learning.

To address these challenges, we propose a novel data synthesis paradigm—MathMixup—which centers on difficulty-controllable hybrid and decomposed strategies to generate high-quality, explicitly graded mathematical reasoning data, thereby both directly improving model capability and laying the groundwork for subsequent curriculum learning. As illustrated in Figure \ref{fig:overview}, the overall process consists of three main stages:
(1) Difficulty-Controllable Question Synthesis: Based on large-scale mathematical datasets, we use BGE embedding \cite{bge_embedding} to construct pairs of similar questions with different difficulty levels, and employ both hybrid generation (to create more challenging new problems) and decomposed generation (to produce problems of intermediate difficulty), thereby achieving fine-grained control over difficulty.
(2) Dataset Construction:  All generated questions undergo automated self-checking and manual screening to ensure semantic clarity and well-graded difficulty. For each synthesized problem, we generate high-quality solutions using QwQ-32B, guided by auxiliary information from similar original questions and their answers. We then apply automated post-processing—including answer format validation and content deduplication—to further ensure the accuracy and consistency of the solutions. This pipeline results in a structured dataset, MathMixupQA, with explicit and reliable difficulty stratification.
(3) Curriculum Learning: During model training, we fully leverage the graded data in MathMixupQA, not only implementing progressive curriculum learning on our own dataset, but also flexibly mixing MathMixupQA data with other public datasets to explore more efficient and diverse training strategies.

Based on the MATH \cite{math} and AMC-AIME training sets, we construct a new dataset, MathMixupQA, featuring controllable difficulty gradients. Experimental results show that curriculum learning based on difficulty-controllable data significantly improves model performance in mathematical reasoning tasks. After curriculum learning SFT on MathMixupQA, Qwen2.5-7B achieves an average score of 47.6\% across seven mathematical benchmarks. Furthermore, when MathMixupQA is mixed with MathFusionQA for curriculum learning, the average performance of Qwen2.5-7B \cite{qwen2025qwen25technicalreport} increases to 52.6\%, setting a new SOTA.

In summary, this work makes three key contributions: \textbf{1)} We propose MathMixup, a framework for difficulty-controllable data synthesis, and construct the MathMixupQA dataset. MathMixup systematically generates mathematical reasoning problems with explicit and controllable difficulty, providing high-quality, well-structured data resources for improving LLM mathematical reasoning. \textbf{2)} We introduce a curriculum learning strategy that leverages the explicit difficulty gradients in MathMixupQA, and supports flexible integration with other datasets. By combining the graded data generated by MathMixupQA with existing datasets, we achieve more efficient and flexible curriculum learning for LLMs. \textbf{3)} Extensive experiments demonstrate the effectiveness and generalizability of our approach. Data synthesis paradigm and curriculum learning strategies of MathMixup significantly enhance the mathematical reasoning abilities of LLMs, outperforming existing baselines and achieving new state-of-the-art results.

\vspace{-5pt}
\section{Related Work}
\vspace{-5pt}
\subsection{Mathematical Reasoning in LLMs}
Large language models (LLMs) have made significant strides in mathematical reasoning, driven by advances in architecture, scaling laws, and large-scale mathematical datasets. Recent models such as DeepSeek R1 \cite{deepseekr1}, OpenAI o3 \cite{gpt4o}, Kimi K1.5 \cite{kimik1_5}, and Qwen3 \cite{qwen3} have achieved higher accuracy and reasoning depth on various benchmarks. However, some LLMs \cite{grattafiori2024llama3herdmodels,jiang2023mistral7b,cai2024internlm2technicalreport} still struggle with complex multi-step reasoning and precise logical deduction. Techniques such as chain-of-thought prompting \cite{xu2025chain,zhang2025enhancing,li2025multimodal}, external tool augmentation \cite{jiang2025deepretrieval,zheng2025deepresearcher,li2024dtllm}, supervised fine-tuning \cite{wen2025light,ma2025cot,tian2025deepdistill}, and reinforcement learning \cite{rstar,xu2025teaching,huang2025hapo,chung2025learning} have been proposed to address these issues. Nevertheless, the lack of high-quality, difficulty-stratified training data remains a key bottleneck. Our work targets this gap by synthesizing training data with controllable difficulty for supervised fine-tuning of LLMs.

\begin{figure*}[htbp!]
  \centering   
  \includegraphics[width=0.95\linewidth]{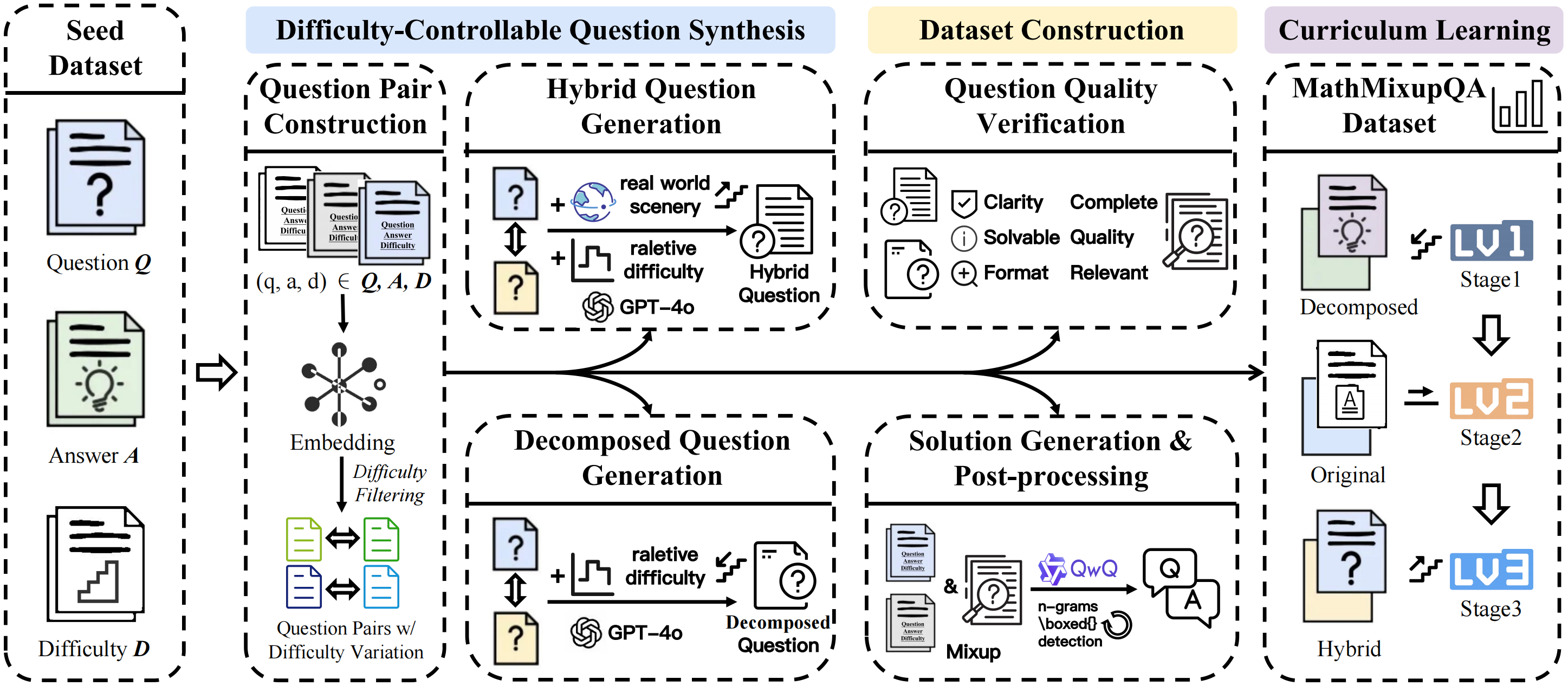}
  \vspace{-8pt}
  \caption{Overview of the MathMixup pipeline. The process includes question pairs construction, question generation (hybrid and decomposed), question verification, solution generation with auxiliary information, and automated post-processing. MathMixup enables the synthesis of high-quality data with controllable difficulty, and supports curriculum learning strategies that further enhance LLM mathematical reasoning performance.}
  \label{fig:overview}
  \vspace{-15pt}
\end{figure*}

\subsection{Data Synthesis for Math Reasoning}
High-quality and diverse datasets \cite{saveliev2025towards,he2025multi,li2024dtllm,li2024dtvlt} are essential for mathematical reasoning in LLMs. Recent works \cite{wizardmath,metamath,mathfusion,mmiqc} have explored automatic problem generation, paraphrasing, and augmentation to enrich data diversity, including template-based and adversarial example construction \cite{orcamath,xwin_math,kpmath,mathscale}. However, most synthetic datasets \cite{refaug,li2025select} still lack fine-grained control over problem complexity and often do not guarantee validity or solvability. Difficulty-controllable synthesis remains largely unexplored. In contrast, our approach enables explicit and fine-grained difficulty control during data synthesis, ensuring each question is both valid and appropriately challenging.

\subsection{Curriculum Learning for Reasoning}
Curriculum learning \cite{shi2025efficient,wang2025learning} improves LLM training efficiency and generalization, especially for tasks with distinct difficulty levels, such as mathematical reasoning. By organizing data according to explicit difficulty levels or conceptual progression \cite{ma2025problem,deng2025boosting,li2025look}, curriculum learning allows models to build foundational knowledge before tackling harder problems. Recent studies \cite{song2025fastcurl,wen2025sari,xia2025improving} have explored manual annotation, automatic difficulty estimation, and adaptive sampling for curriculum design. Yet, reliably constructing controllable difficulty gradations—particularly when blending synthetic data—remains challenging. Our approach addresses this by synthesizing mathematical questions with controllable difficulty, enabling more effective curriculum learning both within our dataset and in combination with others.

\vspace{-5pt}
\section{Method}
\vspace{-5pt}
\subsection{Overview}
The overall MathMixup pipeline is illustrated in Figure~\ref{fig:overview}. Our method consists of three systematic stages: (1) \textbf{Controllable Question Synthesis}, which includes constructing pairs of similar questions and generating new questions with controllable difficulty via hybrid and decomposed strategies; (2) \textbf{Dataset Construction}, involving verifying question quality, solution generation (with auxiliary information) and automated post-processing to build the structured MathMixupQA dataset; and (3) \textbf{Curriculum Learning}, where we leverage the graded data for progressive curriculum learning, both on MathMixupQA alone and in combination with other datasets, enabling more flexible and effective model training.

\subsection{Difficulty-Controllable Question Synthesis}
\subsubsection{Question Pairs Construction}
\label{sec:question-pairs-construction}
To ensure the difficulty and quality of the synthesized data, we use MATH \cite{math} and AMC-AIME as seed training datasets to construct question pairs. The MATH training set contains a total of 7.5K math problems. For the AMC-AIME dataset, we compile AIME problems from 1984-2023 and AMC problems prior to 2023 as training data, totaling 4K math problems.
Let the dataset be denoted as:
\[
\mathcal{D} = \{(q_i, a_i, d_i)\}_{i=1}^N
\]
where \(q_i\) is the \(i\)-th question, \(a_i\) is its answer, and \(d_i\) is the corresponding difficulty level. The difficulty labels \(d_i\) are sourced from official annotations.
For each question \(q_i\), we compute its embedding using BGE embeddings \cite{bge_embedding}:
\[
\mathbf{e}_i = \text{BGE}(q_i)
\]
where \(\mathbf{e}_i \in \mathbb{R}^d\) is the embedding vector for question \(q_i\).
To identify similar question pairs, we calculate the cosine similarity between embeddings:
\[
\text{sim}(q_i, q_j) = \frac{\mathbf{e}_i \cdot \mathbf{e}_j}{\|\mathbf{e}_i\| \|\mathbf{e}_j\|}
\]
We then construct the set of similar question pairs:
\[
\mathcal{P} = \Bigl\{ 
  \bigl((q_i, a_i, d_i), (q_j, a_j, d_j)\bigr) \;\Big|\;
  \begin{aligned}
    &\text{sim}(q_i, q_j) > \tau, \\
    &d_i \neq d_j
  \end{aligned}
\Bigr\}
\]
where \(\tau\) is a predefined similarity threshold, and \(d_i \neq d_j\) ensures that the paired questions have different difficulty levels. By retaining only those pairs with different difficulty levels, we enable better control over the difficulty when synthesizing new questions. This pairing strategy provides the foundation for generating both easier and harder variants of the original problems, supporting explicit difficulty stratification in subsequent synthesis.

\subsubsection{Question Generation}
\label{sec:question-generation}
After obtaining pairs of similar questions, we use GPT-4o \cite{gpt4o} to perform mixup on the two questions in two distinct ways.

For Hybrid Question Generation, the objective is to create a new question that is more difficult than the harder of the two originals. This is achieved by analyzing the mathematical structures, variables, and solution strategies of both problems, then identifying common themes or real-world contexts to construct a unified scenario. The resulting question integrates core elements from both problems and introduces realistic constraints, thereby increasing both difficulty and contextual richness. An example is shown in Case~\ref{trainprompt:hybrid-case}.

\begin{table}[t!]
\begin{trainprompt}{\textit{Hybrid Question Generation}}{hybrid-case}
\small
\textit{\textbf{Original Question 1:} How many integers $n \geq 2$ are there such that whenever $z_1, z_2, \dots, z_n$ are complex numbers with
\[
|z_1| = |z_2| = \dots = |z_n| = 1 \quad \text{and}
\]
\[
z_1 + z_2 + \dots + z_n = 0,
\]
then the numbers $z_1, z_2, \dots, z_n$ are equally spaced on the unit circle in the complex plane?\\
\textbf{Difficulty 1:} 7.0\vspace{6pt}\\
\textbf{Original Question 2:} How many nonzero complex numbers $z$ have the property that $0$, $z$, and $z^3$, when represented by points in the complex plane, are the three distinct vertices of an equilateral triangle?\\
\textbf{Difficulty 2:} 4.0\vspace{6pt}\\
\textbf{Hybrid Question:} A researcher is studying wave patterns generated by buoys anchored at points on the unit circle in the complex plane. The buoys are placed at $n$ points $z_1, z_2, \ldots, z_n$ such that:
\begin{enumerate}
    \item Each $z_k$ is a complex number with $|z_k|=1$.
    \item The buoys are equally spaced around the unit circle (i.e., $z_k = e^{2\pi i k/n}$ for $k=0,1,\ldots,n-1$), so $z_1 + z_2 + \cdots + z_n = 0$.
    \item For each buoy $z_k$, the points $0$, $z_k$, and $z_k^3$ are three distinct vertices of an equilateral triangle in the complex plane.
\end{enumerate}
For how many values of $n \geq 2$ is it possible to place the buoys in this way?\\
\textbf{Difficulty:} 8.0
}
\end{trainprompt}
\vspace{-20pt}
\end{table}

For Decomposed Question Generation, the goal is to synthesize a question whose difficulty lies between that of the two originals. This is done by decomposing and recombining elements from both problems, simplifying complex dependencies while retaining essential concepts. This process produces intermediate-difficulty questions that support a smooth and controllable difficulty progression for curriculum learning. A representative case is shown in Case~\ref{trainprompt:decomposed-case}.

\begin{table}[htbp!]
\begin{trainprompt}{\textit{Decomposed Question Generation}}{decomposed-case}
\small
\textit{\textbf{Original Question 1:} How many integers $n \geq 2$ are there such that whenever $z_1, z_2, \dots, z_n$ are complex numbers with
\[
|z_1| = |z_2| = \dots = |z_n| = 1 \quad \text{and}
\]
\[
z_1 + z_2 + \dots + z_n = 0,
\]
then the numbers $z_1, z_2, \dots, z_n$ are equally spaced on the unit circle in the complex plane?\\
\textbf{Difficulty 1:} 7.0
\vspace{6pt}\\
\textbf{Original Question 2:} How many nonzero complex numbers $z$ have the property that $0$, $z$, and $z^3$, when represented by points in the complex plane, are the three distinct vertices of an equilateral triangle?\\
\textbf{Difficulty 2:} 4.0
\vspace{6pt}\\
\textbf{Decomposed Question:} Let \(z_1, z_2, z_3, z_4\) be complex numbers with \(|z_1| = |z_2| = |z_3| = |z_4| = 1\) and \(z_1 + z_2 + z_3 + z_4 = 0\).  
How many distinct quadruples \((z_1, z_2, z_3, z_4)\) (up to rotation) are there such that the points are not all equally spaced on the unit circle?
\\
\textbf{Difficulty:} 5.0
}
\end{trainprompt}
\vspace{-10pt}
\end{table}

For both generation strategies, prompts guide the creation of logically clear, self-contained problems with unique answers, and include self-check steps to ensure completeness, solvability, and clarity. After hybrid and decomposed question generation, we obtain synthesized data with controllable difficulty compared to the original dataset. The explicit construction of decomposed, original, and hybrid questions provides a three-level difficulty gradient, which is directly leveraged in our curriculum learning framework. For detailed prompts, please refer to Appendix \ref{appendix:prompt}. 

\subsection{Dataset Construction}
\subsubsection{Question Verification}
\label{sec:question-verification}
To ensure the quality and reliability of synthesized questions, we adopt a two-stage verification process, focusing on key dimensions such as clarity, completeness, formatting, relevance, solvability, and logical flow. In the first stage, GPT-4o \cite{gpt4o} performs automated self-checks and corrections, filtering out questions with obvious issues in these aspects. In the second stage, we manually validate a randomly selected 10\% sample of the questions, applying the same criteria. This ensures that each question is clear and understandable, contains all necessary conditions and parameters, uses standard mathematical notation, and is appropriate for the intended academic level. Questions exhibiting serious ambiguity, missing information, logical contradictions, or irrelevant assumptions are flagged as low-quality and excluded. This consistent two-stage process guarantees that all synthesized questions are logically consistent, meaningful, and suitable for inclusion in the dataset, thereby ensuring the overall quality of the difficulty-controlled data.

\vspace{-5pt}
\subsubsection{Solution Generation}
\label{sec:solution-generation}
We used QwQ-32B \cite{qwq32b} to generate solutions for the synthesized data. Existing approaches often rely solely on models like GPT-4o \cite{mathfusion} or use majority voting \cite{guo2025synthetic}, which increases inference costs and may not guarantee correctness, especially for complex problems. In our method, we incorporate original questions and their answers as auxiliary information when generating solutions for new questions. This additional context guides the model to produce answers that are more relevant and better matched to the intended difficulty. Leveraging such mixup-based information reduces noisy outputs and improves efficiency compared to majority voting, which simply selects among independently generated answers.

\subsubsection{Solution Post-processing}
\label{sec:solution-post-processing}
To further ensure the correctness of the generated solutions, we perform post-processing and, if necessary, regeneration. Specifically, we check each solution for the presence of properly formatted \verb|\boxed{}| expressions, ensuring that the answer is complete and not empty. Additionally, we conduct n-gram (2-gram and 3-gram) analysis to detect and prevent repetitive or meaningless content. If any such issues are found, the solution is regenerated. This post-processing step further enhances the quality and clarity of the final solutions in our dataset, making them suitable for robust LLM training.

\subsection{Curriculum Learning}
\label{sec:curriculum-learning}
In the final stage, we train models directly on MathMixupQA dataset and leverage the difficulty-controllable data for curriculum learning (CL) to further enhance the mathematical reasoning abilities of LLMs.
\subsubsection{Curriculum Learning on MathMixupQA}
We design a curriculum learning framework that exploits the explicit difficulty gradation in MathMixupQA. The synthesized questions are organized into several levels (decomposed, original, hybrid) according to their difficulty, and the model is trained progressively from easier to harder problems. This staged process enables the model to build foundational reasoning skills before tackling more complex concepts, providing a data-driven and controllable curriculum learning strategy.
\subsubsection{Blended Curriculum Learning with Other Datasets}
To further improve training flexibility and generalization, we extend our approach by blending MathMixupQA with MathFusionQA datasets. We merge and re-rank questions from both datasets based on their difficulty and categories in Figure \ref{fig:difficulty-score}, constructing a unified curriculum that covers a broader range of problem types and complexities. This blended curriculum allows the model to benefit from more data diversity and explicit difficulty stratification, promoting robust reasoning abilities.

\section{MathMixupQA Dataset}
\subsection{Dataset Statistics}
With MathMixup, we synthesize both Hybrid and Decomposed types of data for the original MATH and AMC-AIME datasets, forming the MathMixupQA dataset. As shown in Table \ref{tab:mathmixupqa_stat}, it contains three difficulty levels—original, hybrid, and decomposed—with 22.5K entries based on MATH and 12K entries based on AMC-AIME, respectively.

\begin{table}[htbp!]
  \vspace{-5pt}
  \centering
  \resizebox{\linewidth}{!}{
    \begin{tabular}{l ccc}
      \toprule
      Dataset & MATH & AMC-AIME & Total \\
      \midrule
      MathMixupQA (\textit{Decomposed}) & 7.5K & 4K & 11.5K \\
      MathMixupQA (\textit{Original}) & 7.5K & 4K & 11.5K \\
      MathMixupQA (\textit{Hybrid}) & 7.5K & 4K & 11.5K \\
      \midrule
      MathMixupQA & 22.5K & 12K & 34.5K \\
      \bottomrule
  \end{tabular}}
  \vspace{-5pt}
  \caption{
    Statistics of the MathMixupQA dataset.
  }
  \label{tab:mathmixupqa_stat}
  \vspace{-15pt}
\end{table}

\subsection{Difficulty Analysis}
\begin{figure}[htbp!]
  \vspace{-10pt}
  \centering   
  \includegraphics[width=\linewidth]{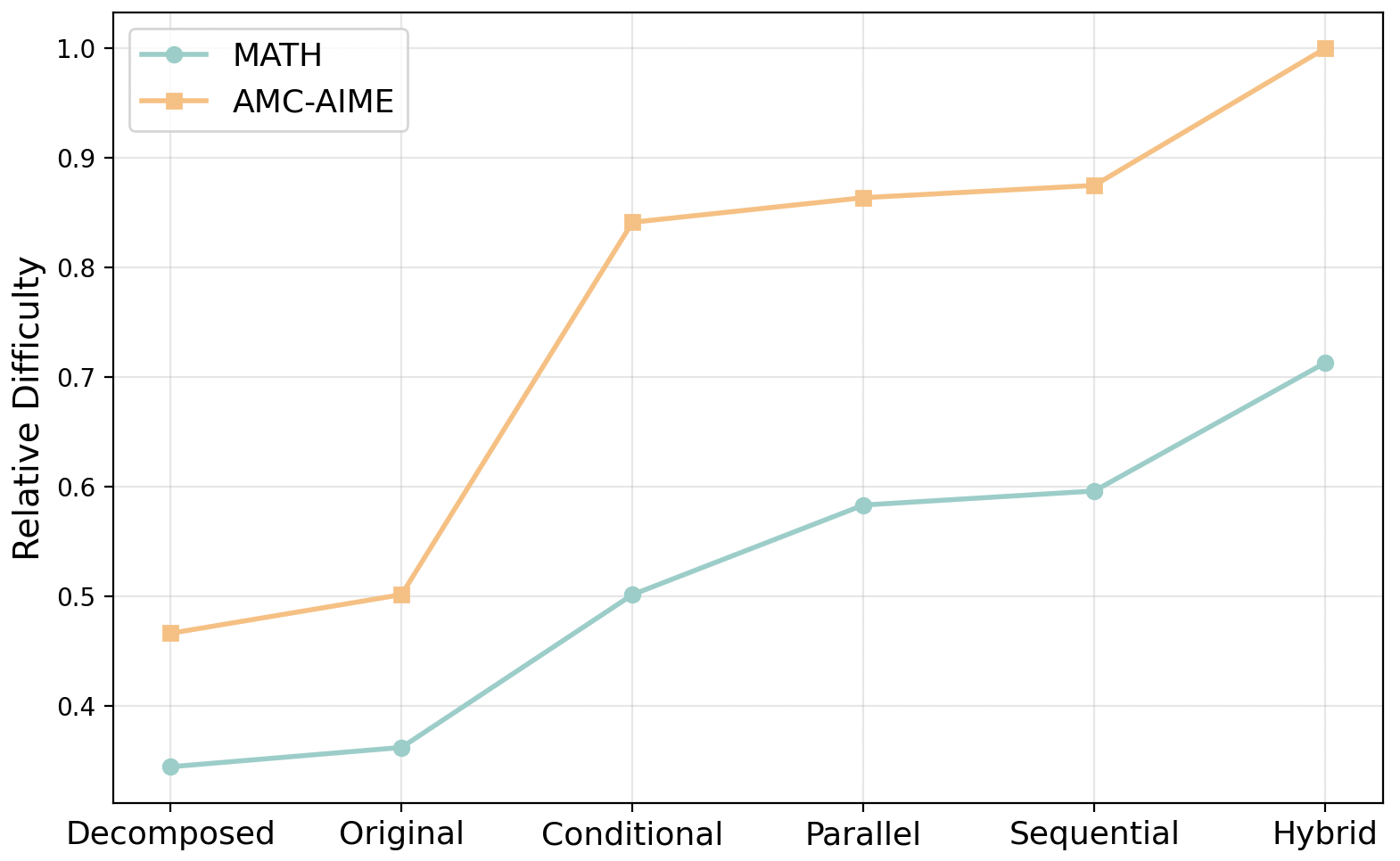}
  \vspace{-20pt}
  \caption{Difficulty scores of different components in the MathMixupQA and MathFusionQA datasets. Note that Conditional, Parallel, and Sequential are synthesis methods proposed in MathFusion.}
  \label{fig:difficulty-score}
  \vspace{-15pt}
\end{figure}

As previously mentioned, MathMixup aims to synthesize data with controllable difficulty, forming three explicit difficulty levels to support curriculum learning and enhance model reasoning. To quantitatively analyze the difficulty distribution, we use the difficulty scoring prompts from DeepScaleR \cite{deepscaler2025} and GPT-4o to assign scores to both MathMixupQA (Decomposed, Original and Hybrid) and MathFusionQA (Conditional, Parallel and Sequential) \cite{mathfusion}. The results, shown in Figure~\ref{fig:difficulty-score}, indicate that the decomposed, original, and hybrid subsets in MathMixupQA form a clear and controllable difficulty gradient, which strongly supports curriculum learning. Notably, the hybrid subset consistently shows higher difficulty scores than MathFusionQA on both MATH and AMC-AIME seeds, further validating the quality and controllability of our data synthesis.

\begin{table*}[htbp!]
    \centering
    \begin{adjustbox}{max width=\textwidth}
        \begin{tabular}{lcccccccccc}
          \toprule
          \multicolumn{1}{c}{\multirow{1}{*}{Method}} & Seed & \# Samples & AIME25 & AIME24 & OlympiadBench & AMC23 & MATH500 & Minerva Math & GSM8K & AVG \\
          \midrule
          \multicolumn{11}{c}{\textbf{Qwen2.5-7B}} \\
          \midrule
          MathFusion$^\dagger$ & \multirow{3}{*}{MATH} & 22.5K & 27.19 & 26.77 & 33.93 & 44.58 & 74.60 & 25.00 & 87.11 & 45.60 \\
          MathMixup &  & 22.5K & 28.13 & 28.33 & 35.85 & 45.78 & 74.20 & 25.37 & 86.58 & 46.32 \\
          MathMixup-CL &  & 7.5K*3 & \textbf{28.33} & \textbf{28.75} & \textbf{36.74} & \textbf{46.99} & \textbf{76.80} & \textbf{26.47} & \textbf{87.49} & \textbf{47.37} \\
          \hdashline
          MathFusion$^\dagger$ & \multirow{3}{*}{AMC-AIME} & 12K & 28.23 & 28.65 & 34.67 & \textbf{45.78}	& 71.40	& 23.16	& 86.28 & 45.45 \\
          MathMixup &  & 12K & 27.81 & 30.94 & 35.70 & \textbf{45.78} &73.80 & 26.10 & 87.49 & 46.80 \\
          MathMixup-CL &  & 4K*3 & \textbf{28.96} & \textbf{32.50} & \textbf{36.00} & 44.58 & \textbf{75.40} & \textbf{27.57} & \textbf{88.17} & \textbf{47.60} \\
          \midrule
          \multicolumn{11}{c}{\textbf{InternLM2.5-7B}} \\
          \midrule
          MathFusion$^\dagger$ & \multirow{3}{*}{MATH} & 22.5K & 9.27	& 6.35 & 18.52	& \textbf{24.10} & 53.80 & 8.82	& 74.37 & 27.89 \\
          MathMixup &  & 22.5K & 8.54	& \textbf{7.50} & 18.81 & \textbf{24.10} & \textbf{56.80}	& 10.29	& 75.06 & 28.73 \\
          MathMixup-CL &  & 7.5K*3 & \textbf{11.25} & 6.88	& \textbf{19.70}	& \textbf{24.10}	& 56.20	& \textbf{12.13}	& \textbf{75.28} & \textbf{29.36} \\
          \hdashline
          MathFusion$^\dagger$ & \multirow{3}{*}{AMC-AIME} & 12K & 11.15 & 7.50	& 15.41	& 18.07	& 45.20 & 10.66 & 70.36 & 25.48  \\
          MathMixup &  & 12K & \textbf{13.54} & 7.60 & \textbf{17.78} & 19.28 & 46.40 & 9.56	& 70.74 & 26.41  \\
          MathMixup-CL &  & 4K*3 & 12.29 & \textbf{9.48} & 17.04 & \textbf{21.69} & \textbf{48.00} & \textbf{11.76} & \textbf{75.80} & \textbf{28.01} \\
          \midrule
          \multicolumn{11}{c}{\textbf{LLaMA3.1-8B}} \\
          \midrule
          MathFusion$^\dagger$ & \multirow{3}{*}{MATH} & 22.5K & 10.00 & 5.63 & 17.33 & 19.28 & 48.40 & 10.29 & 72.71 & 26.23 \\
          MathMixup &  & 22.5K & \textbf{11.25} & \textbf{8.13}	& 17.04	& 21.69	& \textbf{52.80} & 11.03 & 73.01 & 27.85 \\
          MathMixup-CL &  & 7.5K*3 & 10.10	& 6.15 & \textbf{19.26}	& \textbf{26.51}	& 52.60 & \textbf{11.40} & \textbf{75.89} & \textbf{28.84} \\
          \hdashline
          MathFusion$^\dagger$ & \multirow{3}{*}{AMC-AIME} & 12K & 11.67 & 6.67	& 15.11	& 14.46	& 42.20 & 11.40	& 68.23 & 24.25 \\
          MathMixup &  & 12K & 13.13 & 8.13 & \textbf{17.78} & 16.87 & \textbf{43.20} & 9.56	& 73.01 & 25.95 \\
          MathMixup-CL &  & 4K*3 & \textbf{13.75} & \textbf{9.27} & 16.89	& \textbf{18.07}	& \textbf{43.20}	& \textbf{12.50}	& \textbf{73.16} & \textbf{26.69} \\
          \bottomrule
        \end{tabular}
    \end{adjustbox}
    \vspace{-8pt}
    \caption{Performance comparison on mathematical benchmarks including AIME25, AIME24, OlympiadBench, AMC23, MATH500, Minerva Math and GSM8K. 
    The table is organized by the base model and seed datasets. Baseline labeled with $^\dagger$, which is our own runs. The best results are highlighted in bold.}
    \label{tab:main_results}
    \vspace{-15pt}
\end{table*}

\section{Experiment}
\subsection{Implementation Details}

\subsubsection{Data Synthesis.} We use GPT-4o \cite{gpt4o} to synthesize new questions from similar question pairs, and use QwQ-32B \cite{qwq32b} to generate the long CoT responses. We empirically set the similarity threshold $\tau$ in the range of 0.75 to 0.9.

\subsubsection{Training.} For supervised fine-tuning (SFT), we use 360-LLaMA-Factory \cite{360-llama-factory} to train each model for 3 epochs, with a cosine learning rate schedule, 1\% warm-up, maximum learning rate 1e-5, and sequence length 32,768 tokens. Experiments are conducted on Qwen2.5-7B \cite{qwen2025qwen25technicalreport}, LLaMA3.1-8B \cite{grattafiori2024llama3herdmodels}, and InternLM2.5-7B \cite{cai2024internlm2technicalreport}. Each model is trained directly on MathMixupQA based on different seed datasets and we use a three-stage SFT process aligned with the decomposed, original, and hybrid subsets for curriculum learning.
To ensure fair comparison, we also reproduce MathFusion using the same data synthesis protocol as our baseline. Further details are provided in the Appendix \ref{appendix:reproduction}.
To evaluate generalizability, we conduct additional experiments by blending MathMixupQA with MathFusionQA, constructing a mixed training set. We perform both standard SFT (Blending) and curriculum learning SFT (Blending-CL), with curriculum stages determined by the difficulty gradients in Figure~\ref{fig:difficulty-score}. This setup allows us to assess the benefits of integrating our difficulty-controllable data with existing high-quality datasets. Note that we group every two consecutive difficulty levels into one training stage.

\subsection{Evaluation and Metrics}
We evaluate the fine-tuned models on AMC2023, MATH \cite{math}, Minerva Math \cite{minerva}, OlympiadBench \cite{olympiadbench}, GSM8K \cite{gsm8k}, AIME2024, and AIME2025, using the Light-R1 framework \cite{light-r1}. For the first five benchmarks, we report avg@1. Due to the smaller size of AIME2024 and AIME2025, we use the average of 32 random samplings (Avg@32) for more stable results. For detailed hyperparameter settings, please refer to the Appendix \ref{appendix:hyperparameter}.

\subsubsection{SFT on MathMixupQA}
As shown in Table~\ref{tab:main_results}, MathMixup consistently outperforms the MathFusion baseline across all models and both seed datasets under the same data volume. For example, Qwen2.5-7B fine-tuned on MathMixupQA based on AMC-AIME achieves 46.80\% average accuracy, surpassing MathFusionQA’s 45.45\%. This trend holds across individual benchmarks and other models. Applying curriculum learning (MathMixup-CL) further improves performance for all models, with Qwen2.5-7B reaching 47.60\% and notable gains on challenging benchmarks such as OlympiadBench.
We also observe that models with higher baseline performance, such as Qwen2.5-7B, benefit more from increased data diversity and curriculum learning, especially when trained with more challenging data. In contrast, models with lower baseline abilities, like LLaMA3.1-8B, gain less from high-difficulty data without sufficient curriculum support. These results highlight the importance of tailoring data difficulty and curriculum design to match model capacity.
Overall, MathMixup provides higher-quality, difficulty-controllable training data than existing baselines. The improvements across models and benchmarks demonstrate the robustness and generalizability of our approach.

\subsubsection{SFT on Blending Datasets}
\begin{figure}[htbp!]
  \vspace{-5pt}
  \centering   
  \includegraphics[width=\linewidth]{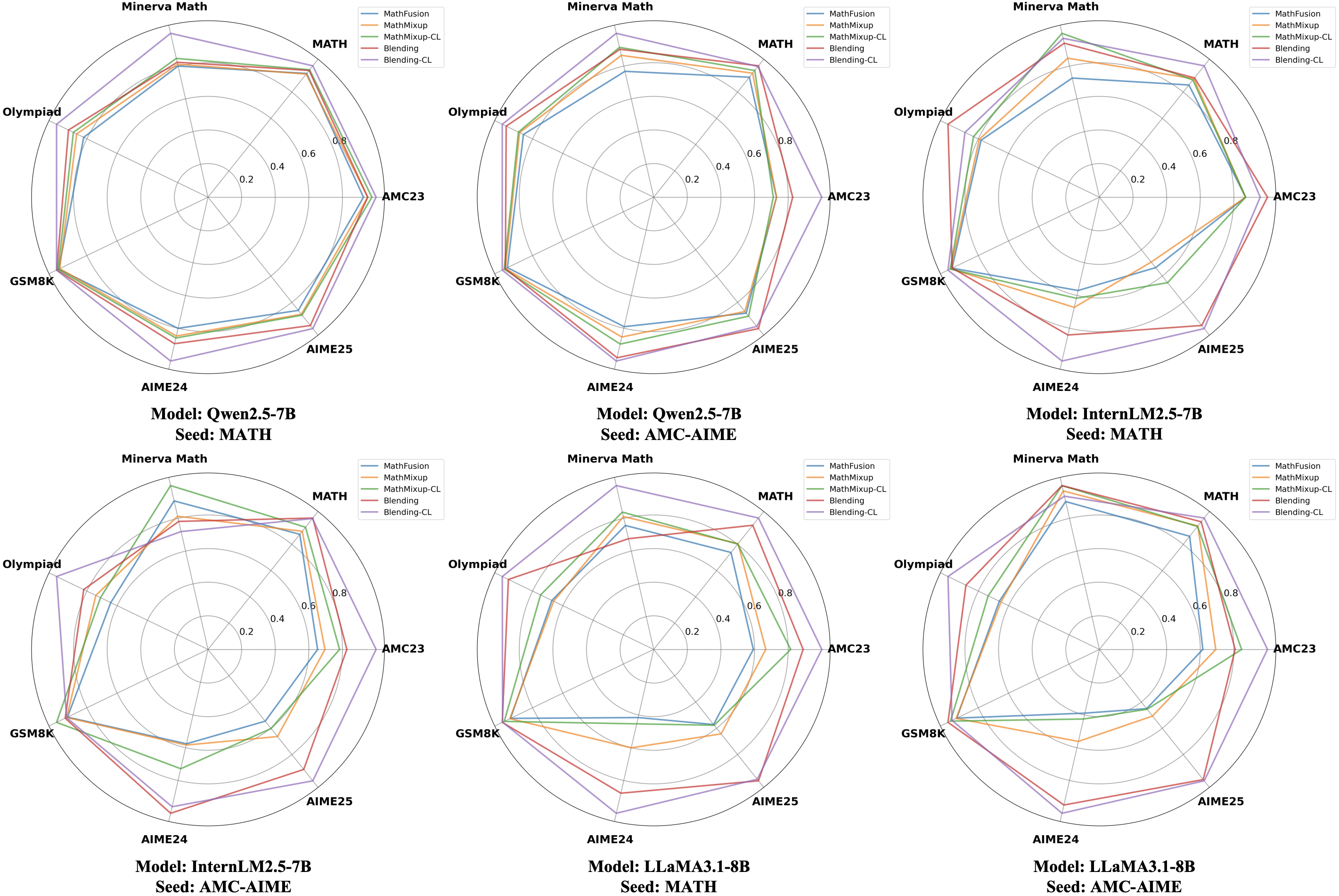}
  \vspace{-15pt}
  \caption{Normalized performance on seven mathematical reasoning benchmarks for three LLMs using different data synthesis and training strategies. “Blending” denotes SFT on the mixed dataset of MathMixupQA and MathFusionQA, while “Blending-CL” denotes curriculum learning SFT on the same mixture.}
  \label{fig:combination_results}
  \vspace{-15pt}
\end{figure}

As illustrated in Figure~\ref{fig:combination_results}, blending MathMixupQA and MathFusionQA (Blending) further improves model performance compared to using either dataset alone. This effect is amplified with curriculum learning (Blending-CL), as shown by the outward expansion of the radar plots. To account for the large differences in absolute scores across benchmarks, we normalize the results. The gains are consistent across both easier (e.g., GSM8K, MATH500) and more challenging (e.g., AIME24, OlympiadBench) benchmarks. These results confirm that combining difficulty-controllable data with existing high-quality datasets, together with curriculum learning, can fully unlock the potential of LLMs for mathematical reasoning, achieving new state-of-the-art performance. Detailed numerical results are provided in the Appendix \ref{appendix:results}.

\subsection{Ablation Study}
\subsubsection{Training Data Components}
As shown in Table~\ref{tab:data_components}, both hybrid and decomposed data play critical roles in enhancing model reasoning performance. Consistent findings emerge across the two seed datasets (MATH and AMC-AIME): By comparing combinations 1 vs. 3 and 2 vs. 3, we observe that decomposed and hybrid data contribute more significantly to improving reasoning abilities than original data. Furthermore, combination 3 (integrating both decomposed and hybrid data) achieves the optimal results, validating the value of these two data components. Note that we show the average accuracy.

\begin{table}[htbp!]
\vspace{-5pt}
\centering
\resizebox{0.95\linewidth}{!}{
\begin{tabular}{c|c|ccc|c}
\toprule
Seed & \# & Decomposed & Original & Hybrid & AVG \\
\midrule
\multirow{3}{*}{MATH}
      & 1 &       & \cmark & \cmark & 44.37 \\
      & 2 & \cmark & \cmark &       & 43.36 \\
      & 3 & \cmark &       & \cmark & 45.67 \\
\midrule
\multirow{3}{*}{AMC-AIME}
      & 1 &       & \cmark & \cmark & 44.83 \\
      & 2 & \cmark & \cmark &       & 43.59 \\
      & 3 & \cmark &       & \cmark & 45.65 \\
\bottomrule
\end{tabular}
}
\vspace{-5pt}
\caption{Effect of data components on Qwen2.5-7B.}
\label{tab:data_components}
\vspace{-15pt}
\end{table}

\subsubsection{Curriculum Leaning Stage Order}
As shown in Table~\ref{tab:data_curriculum_learning}, curriculum learning that starts with decomposed data followed by hybrid data (Decomposed+Hybrid) consistently achieves the highest accuracy across seven challenging benchmarks. Using original data as the first stage yields lower accuracy, and Decomposed+Original performs worst. These findings demonstrate that introducing decomposed data early in the curriculum better prepares the model for complex questions, highlighting the importance of explicit difficulty control in curriculum design. We provide more results of ablation studies in Appendix \ref{appendix:ablation}.

\begin{table}[htbp!]
\vspace{-5pt}
\centering
\resizebox{0.8\linewidth}{!}{
\begin{tabular}{c|cc|c}
\toprule
Seed & Stage1 Data & Stage2 Data & AVG \\
\midrule
\multirow{3}{*}{MATH}
& Decomposed & Original & 43.24 \\
& Original  & Hybrid   & 46.19 \\
& Decomposed & Hybrid   & 46.88 \\
\midrule
\multirow{3}{*}{AMC-AIME}
& Decomposed & Original & 43.90 \\
& Original  & Hybrid   & 46.72 \\
& Decomposed & Hybrid   & 47.26 \\
\bottomrule
\end{tabular}
}
\vspace{-5pt}
\caption{Effect of curriculum learning stage order on Qwen2.5-7B across seven benchmarks.}
\label{tab:data_curriculum_learning}
\vspace{-15pt}
\end{table}

\subsubsection{Effectiveness of Solution Generation}
\begin{figure}[htbp!]
  \centering   
  \vspace{-5pt}
  \includegraphics[width=\linewidth]{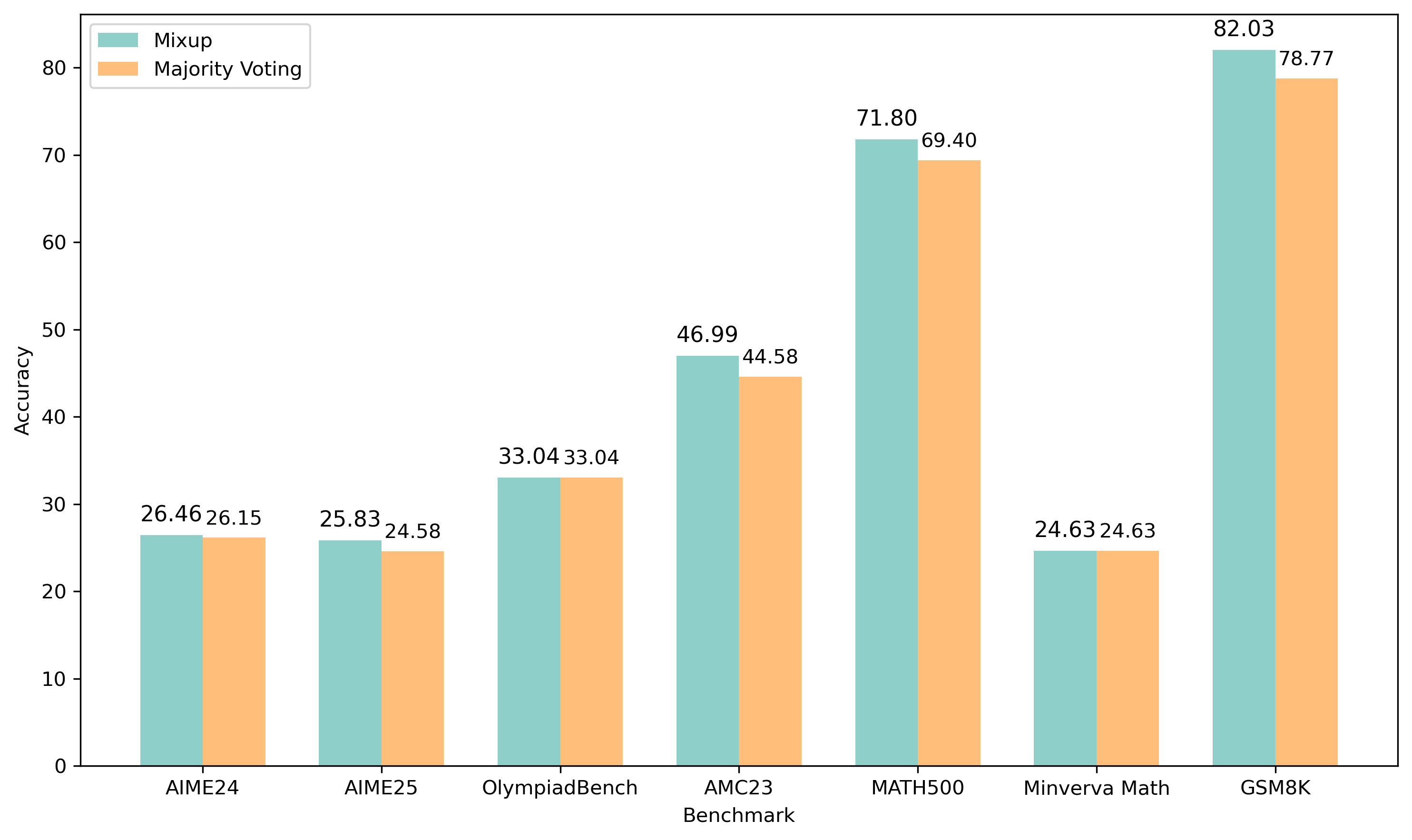}
  \vspace{-25pt}
  \caption{Effectiveness and efficiency of solution generation in MathMixup.}
\label{fig:mixup}
\vspace{-15pt}
\end{figure}

As illustrated in Figure~\ref{fig:mixup}, the solution generation strategy consistently outperforms majority voting with 16 candidate solutions across benchmarks. While majority voting is commonly used to filter noisy solutions, it is computationally expensive and does not always guarantee correctness. By incorporating relevant original questions and answers as auxiliary information, we yield higher-quality and more reliable solutions, while also reducing inference costs. These improvements further strengthen the overall effectiveness of the MathMixup.

\section{Conclusion}
We propose MathMixup, a novel method for boosting LLM mathematical reasoning with difficulty-controllable data synthesis and curriculum learning. By explicitly generating and organizing problems with controllable difficulty levels, MathMixup enables the construction of the MathMixupQA dataset and supports the design of controllable, data-driven curricula for LLMs. Extensive experiments show that both MathMixupQA and curriculum learning strategies significantly boost model performance, achieving new SOTA results.

\section*{Limitations}
Although MathMixup provides a principled way to synthesize difficulty-controllable mathematical reasoning data, several constraints remain. The automatic generation and validation process cannot fully eliminate subtle errors in problem statements or solutions. Due to the high cost of manual review, only about 10\% of the synthesized problems were randomly inspected, which may leave undetected issues in the remaining data. Moreover, our study primarily investigates two-level fusion strategies, leaving richer compositions of multiple problems and alternative retrieval mechanisms largely unexplored. Finally, the constructed MathMixupQA dataset, though useful for curriculum learning, is modest in size compared with broader mathematical corpora, and its domain coverage is not yet exhaustive.

\bibliography{custom}

\clearpage
\appendix

\renewcommand{\thetable}{A\arabic{table}}
\renewcommand{\thefigure}{A\arabic{figure}}
\setcounter{figure}{0}  
\setcounter{table}{0}  

\section*{Appendix}
\section{Prompts}
\label{appendix:prompt}
We present the key steps of the prompts used for Hybrid Generation in Prompt~\ref{trainprompt:hybrid}, Decomposed Generation in Prompt~\ref{trainprompt:decomposed} and Question Verification in Prompt~\ref{trainprompt:verify}, which are partially derived from MathFusion.

\begin{table*}[t!]
\begin{trainprompt}{\textit{Hybrid Question Generation}}{hybrid}
\allowdisplaybreaks
\small
\textit{\#\#\# Role: Hybrid Scenario Problem Architect\\\\
\#\#\#\# Profile\\
You are an expert in designing advanced, real-world mathematical problems. Your task is to create a hybrid real-world scenario that seamlessly blends "\#Problem 1\#" and "\#Problem 2\#", ensuring their essential mathematical elements are preserved and deeply intertwined, leading to a single, culminating challenge. The new problem must be self-contained and must not reference or mention Problem 1 or Problem 2.\\\\
\#\#\#\# Guidelines\\
Step 1: Carefully analyze the mathematical structures, variables, solution strategies, standard answers, and difficulty levels of both problems, as well as their possible real-world interpretations.\\\\
Step 2: Identify common themes, physical principles, or practical contexts that can naturally link the two problems together.\\\\
Step 3: Construct a hybrid scenario where these themes converge, introducing realistic constraints and details that bind the mathematical frameworks of both problems into a coherent, practical setting.\\\\
Step 4: Design a sophisticated, unified problem statement where solving the underlying mathematical challenges from both original problems is necessary to resolve the real-world scenario.\\\\
Step 5: When constructing the new problem, take into account the standard answers and difficulty levels of both original problems. Avoid making the new problem’s answer a simple sum, product, or direct reuse of the original answers. Ensure the new problem’s difficulty is at least as high as the more difficult of the two original problems, and ideally, it should present a new layer of challenge or synthesis.\\\\
Step 6: The new problem must be a single, standalone question with only one main objective and must not be split into multiple subproblems, parts, or steps. The answer should be unique and clearly defined. Do not require the solver to provide multiple separate answers, perform open-ended analysis, or combine results from different objectives.\\\\
Self-Check and Correction:\\\\
After constructing the hybrid problem, carefully review it by answering the following:\\
- Does the problem statement include all necessary information and constraints?\\
- Does the scenario contain any irrelevant or unnecessary details?\\
- Is the problem well-posed and solvable (i.e., is there at least one solution, and is the solution unique or well-defined)?\\
- Does the problem require only one specific answer, and is the objective clearly stated as a single maximization, minimization, or unique solution?\\\\
If any issues are found, revise and improve the problem statement to ensure it is complete, relevant, solvable, and meets the single-objective requirement.\\\\
Only output the final, corrected problem statement. Do not include the self-check process or any commentary in your final output.\\\\
Important:\\
- Do not mention or allude to Problem 1, Problem 2, or their answers/difficulty in the new problem.\\
- The final problem must have only one main objective and require only one answer.\\
- The new problem’s objective must be clearly defined and not require separate maximization and minimization.\\
- Do not split the new problem into multiple subproblems, parts, or steps. Do not require multiple distinct answers or open-ended analysis. It must be a single, unified question with a single, well-defined answer.\\
- After outputting the new problem statement, do not add any further explanation, commentary, or additional information. End your output immediately after the new problem statement.\\\\
\#\#\#\# Output Format\\
Please reply strictly in the following format:\\\\
\#Core Elements\#:\\
(Briefly list the main mathematical concepts, variables, or techniques from both problems that will be integrated.)\\\\
\#Scenario Integration\#:\\
(Describe how the real-world scenario is constructed to blend the mathematical elements of both problems, and how the scenario ensures the necessity of resolving both underlying mathematical challenges.)\\\\
\#New Problem\#:\\
(Present the fully integrated, standalone, and self-checked real-world problem statement with a single, clearly defined objective and a unique answer. Do not split it into multiple subproblems or parts. Do not require multiple answers or open-ended analysis. End your output here.)
}
\end{trainprompt}
\vspace{-15pt}
\end{table*}

\section{MathFusion Reproduction}
\label{appendix:reproduction}
Since MathFusion does not perform data synthesis on the AMC-AIME seed dataset and the GSM8K seed dataset it uses is relatively simple, we reproduce MathFusion using the MATH and AMC-AIME seed datasets to ensure a fair comparison. We adopt the three synthesis prompts provided in the original paper (Conditional, Sequential, and Parallel), without controlling for difficulty. Apart from the prompt types and the lack of difficulty control, all other procedures remain consistent with those used in MathMixup to guarantee comparability between the two methods. Additionally, during training, we downsample the synthetic data generated by our MathFusion reproduction to match the amount of synthetic data used in our own method.

\begin{table*}[t!]
\begin{trainprompt}{\textit{Decomposed Question Generation}}{decomposed}
\small
\textit{\#\#\# Role: Advanced Mathematical Problem Decomposer\\\\
\#\#\#\# Profile:\\
You excel at decomposing advanced mathematical problems, transforming originally complex problems into easier-to-master questions, while retaining their educational value and challenge. Your task is to analyze a combination of two similar but differently difficult problems, namely "\#Problem 1\#" and "\#Problem 2\#". Based on the lower difficulty problem, simplify the higher difficulty problem by reducing its complex parts while retaining the essential concepts. The result should be a new problem that still offers learning and thought value. The new problem must be self-contained and should provide clear steps to reach the solution, without mentioning or referencing Problem 1 or Problem 2.\\\\
\#\#\#\# Guidelines:\\
Step 1: Carefully analyze the mathematical structures, variables, solution strategies, and difficulty levels of the two challenging problems provided, especially the problem with high difficulty.\\\\
Step 2: Identify opportunities to simplify the problem:\\
- Consider the similarities between the high and low difficulty problems and simplify complex interdependencies between variables to more direct relationships.\\
- Break the problem down into simple components focused on key concepts.\\
- Design the new problem so that one logical method leads to the solution.\\
- Ensure it maintains a logical sequence, clarity, and avoids any confusion. If new variables or conditions are introduced, make their determination clear and unambiguous.\\\\
Step 3: Identify common themes of the two problems. After that, consider the way to discompose the two problems and design a new problem.
When designing the new problem, consider the standard answers and difficulty levels of the two original problems. Avoid making the answer a simple subtraction, division, or direct reuse of the original answers. Ensure that the difficulty of the new problem is at least as high as the simpler of the two original problems. Ideally, it should present a clearer problem statement.\\\\
Step 4: Craft a coherent problem statement that encompasses simplified elements derived from the original complex problems, ensuring all necessary information is provided for solving the problem. Design a clear problem statement fully considering elements from both problems, ensuring all interdependencies are clear and necessary for the solution. The new problem must be a single, standalone question with only one main objective and must not be divided into multiple subproblems, parts, or steps. The answer should be unique and clearly defined.\\\\
Self-check and Revision:\\\\
After creating the new problem, review it by answering the following questions:\\
- Does the problem statement include all necessary information and constraints?\\
- Is the problem well-posed and solvable (i.e., is there at least one solution, and is the solution clear)? If any issues are found, revise and improve the problem statement to ensure it is complete, relevant, and solvable.\\
- Adjust parts that could lead to misunderstandings or ambiguity.\\
- Thoroughly review the  output to spot any unreasonable elements, ensuring it is an easier version of "\#Problem 1\#", but has higher difficulty than "\#Problem 2\#".\\\\
Important:\\
- Do not mention or imply specific details or answers of the original challenging problems in the new problem.\\
- The final problem must have only one main objective and require only one answer.\\
- Ensure the new problem remains unified and is not split into multiple subproblems or parts.\\
- After presenting the new problem statement, do not include further explanation, commentary, or additional information. End your output immediately after the new problem statement.\\
- For a given set of problem1 and problem2, output only one decomposed problem.\\
- The generated problem must not be presented in a multiple-choice format.\\
- Conclude the output immediately after presenting the new statement.\\\\
\#\#\#\# Output Format:\\
Please respond strictly in the following format, do not include any content from the input.:\\\\
\#Core Elements\#:\\
(Briefly list the key mathematical concepts, variables, or techniques that will be simplified and integrated into the new problem.)\\\\
\#Simplification Strategy\#:\\
(Explain how simplification is achieved by reducing complexity and focusing on key concepts. Describe how the new problem's difficulty compares to the original challenging problems.)\\\\
\#New Problem\#:\\
(Present the new, simplified problem statement, ensuring it has a single objective, is clear, and solvable. End your output here.)
}
\end{trainprompt}
\vspace{-15pt}
\end{table*}

\begin{table*}[t!]
\begin{trainprompt}{\textit{Question Verification}}{verify}
\small
\textit{Your task is to act as an impartial judge to evaluate the statement completeness, correctness, and overall quality of a synthesized math problem according to the following rules:\\\\
1. Clarity: Is the problem statement mostly clear and understandable, even if some wording is informal or not perfectly concise?\\\\
2. Completeness: Are the main conditions, constraints, and parameters provided, so that a reasonably skilled student could attempt the problem? Minor omissions or the need for standard mathematical assumptions are acceptable.\\\\
3. Formatting: Is the problem readable and uses standard mathematical notation, even if there are minor formatting or typographical inconsistencies?\\\\
4. Relevance: Is the problem generally relevant and appropriate for the intended academic level and topic?\\\\
5. Solvability: Is the problem likely solvable by standard mathematical methods, with at least one reasonable solution? (It is acceptable if the solution is not unique, as long as the problem is meaningful.)\\\\
6. Logical Flow: Is the problem statement overall logical and consistent, and free from major irrelevant or confusing information? Minor awkwardness or redundancy is acceptable.
}
\end{trainprompt}
\end{table*}

\section{Hyperparameter Settings}
\label{appendix:hyperparameter}
\subsection{Data Synthesis.} For data synthesis, we set GPT-4o’s temperature to 0.7 and the maximum generation length to 4096 tokens. For QwQ-32B, we use the official recommended hyperparameters: temperature 0.6, TopP 0.95, MinP 0, TopK 40, and no repetition penalty. The maximum response length for QwQ-32B is set to 32,768 tokens.

\subsection{Training.} For supervised fine-tuning (SFT), we train all parameters on the MathMixupQA dataset using DeepSpeed ZeRO Stage 3 for efficient scaling. The effective batch size per device is 1, with gradient accumulation over 4 steps, and sequence parallelism of 8. We use the Adam optimizer ($\beta_1=0.9$, $\beta_2=0.95$, $\epsilon=1\times10^{-8}$), weight decay 0.1, and gradient clipping at norm 1.0. Training uses bfloat16, FlashAttention v2, and gradient checkpointing to save memory. All experiments use a fixed random seed for reproducibility.

\subsection{Evaluation.} For the standard avg@1 setting, each model is evaluated with temperature 0 and one sample per input. The model generates responses up to 32,768 tokens, with a batch size of 2048 and top-p 0.95. For avg@32, we follow SimpleRL-Zoo’s temperature of 1.0 and generate 32 samples per input, keeping other parameters unchanged.

\section{Dataset Statistics and Comparison}
\label{appendix:statistics}
In Table~\ref{tab:datasets_stat}, we compare MathMixupQA with previous mathematical datasets. Despite its relatively moderate scale, our experiments demonstrate that MathMixupQA achieves superior performance under the same data volume, surpassing the state-of-the-art (SOTA) MathFusion method. Furthermore, the explicit difficulty control within each component of MathMixupQA enables effective curriculum learning, which further improves the model's mathematical reasoning performance.

\begin{table*}[t!]
  \centering
  \resizebox{0.6\linewidth}{!}{
    \begin{tabular}{lcc}
      \toprule
      \textbf{Dataset} & \textbf{Total Samples} & \textbf{Difficulty Levels} \\
      \midrule
      WizardMath & 96K & -- \\
      MetaMathQA & 395K & -- \\
      MMIQC & 2294K & -- \\
      Orca-Math & 200K & -- \\
      Xwin-Math-V1.1 & 1440K & -- \\
      KPMath-Plus & 1576K & -- \\
      MathScaleQA & 2021K & -- \\
      DART-Math-Uniform & 591K & -- \\
      DART-Math-Hard & 585K & -- \\
      RefAug & 30K & -- \\
      MathFusionQA & 60K & -- \\
      \midrule
      MathMixupQA-M (MATH) & 22.5K & \textbf{Explicit: 3} \\
      MathMixupQA-A (AMC-AIME) & 12K & \textbf{Explicit: 3} \\
      \bottomrule
    \end{tabular}
  }
  \caption{
    Comparison between MathMixupQA and previous mathematical datasets. MathMixupQA is the only dataset that explicitly controls difficulty levels during data synthesis, resulting in three well-defined difficulty gradients (Original, Hybrid, Decomposed) for both MATH and AMC-AIME seeds.
  }
  \label{tab:datasets_stat}
\end{table*}

\section{Results of SFT on Blending Datasets}
\label{appendix:results}
Tables \ref{tab:math_seed_results} and \ref{tab:amc_aime_seed_results} provide the detailed numerical results for three representative models (Qwen2.5-7B, InternLM2.5-7B, and LLaMA3.1-8B) on seven mathematical benchmarks, using the MathMixupQA dataset with both Blending and Blending-CL settings. Results are reported separately for the MATH and AMC-AIME seed datasets. The raw scores presented in these tables correspond to the original (unnormalized) benchmark results, which serve as the basis for the normalized radar charts shown in Figure 4 of the main text.

\section{Method}
\label{appendix:algorithm}
The detailed pseudocode of the MathMixup method is provided in Algorithm \ref{alg:mathmixup}.

\begin{algorithm*}[t!]
\caption{MathMixup Method}
\label{alg:mathmixup}
\begin{algorithmic}[1]
\REQUIRE Seed datasets $\mathcal{D}_{\text{seed}}$, original data $\mathcal{D}_{\text{orig}}$, (optional) external data $\mathcal{D}_{\text{ext}}$

\STATE \textbf{// 1. Difficulty-Controllable Question Synthesis}
\FORALL{$\mathcal{D}$ in $\mathcal{D}_{\text{seed}}$}
    \FORALL{$(q_i, a_i, d_i) \in \mathcal{D}$}
        \STATE $\mathbf{e}_i = \mathrm{BGE}(q_i)$
        \FORALL{$(q_j, a_j, d_j) \in \mathcal{D},\, j \neq i$}
            \STATE $\mathbf{e}_j = \mathrm{BGE}(q_j)$; $s_{ij} = \cos(\mathbf{e}_i, \mathbf{e}_j)$
            \IF{$s_{ij} > \tau$ \AND $d_i \neq d_j$}
                \STATE Add $((q_i, a_i, d_i), (q_j, a_j, d_j))$ to $\mathcal{P}$
            \ENDIF
        \ENDFOR
    \ENDFOR
\ENDFOR
\FORALL{pair $((q_1, a_1, d_1), (q_2, a_2, d_2)) \in \mathcal{P}$}
    \STATE Generate and verify $q^{\text{hyb}}, q^{\text{dec}}$; add to $\mathcal{D}_{\text{syn}}$
\ENDFOR

\STATE \textbf{// 2. Dataset Construction}
\FORALL{$q \in \mathcal{D}_{\text{syn}}$}
    \STATE Find auxiliary set $\mathcal{A}_q$
    \STATE Generate solution $a = \mathrm{QwQ}(q, \mathcal{A}_q)$
    \STATE Post-process and add $(q, a)$ to $\mathcal{D}_{\text{MathMixupQA}}$
\ENDFOR

\STATE \textbf{// 3. Curriculum Learning}
\STATE Split $\mathcal{D}_{\text{MathMixupQA}}$ by difficulty
\IF{$\mathcal{D}_{\text{ext}} \neq \emptyset$}
    \STATE $\mathcal{D}_{\text{blend}} = \mathrm{BlendAndRank}(\mathcal{D}_{\text{MMQA}}, \mathcal{D}_{\text{ext}})$
    \STATE split by difficulty
\ENDIF
\FORALL{curriculum stage $\mathcal{D}_{\text{stage}}$}
    \STATE Fine-tune LLM on $\mathcal{D}_{\text{stage}}$
\ENDFOR
\end{algorithmic}
\end{algorithm*}

\section{Ablation Study}
\label{appendix:ablation}
We provide the evaluation results of the fine-tuned Qwen2.5-7B on seven benchmarks in Tables \ref{tab:data_components} and \ref{tab:data_curriculum_learning}.

\begin{table*}[t!]
\centering
\resizebox{0.93\textwidth}{!}{
\begin{tabular}{c|c|cccccccccc}
\toprule
            Model & Type & AMC23 & MATH & Minerva Math & OlympiadBench & GSM8K & AIME24 & AIME25 \\
            \midrule
            \multirow{2}{*}{Qwen2.5-7B}
            & Blend & 45.78 & 76.40 & 25.74 & 38.07 & 88.32 & 29.90 & 30.83 \\
            & Blend-CL & \textbf{48.19} & \textbf{79.20} & \textbf{31.25} & \textbf{41.33} & \textbf{88.40} & \textbf{33.44} & \textbf{31.56} \\
            \midrule
            \multirow{2}{*}{InternLM2.5-7B}
            & Blend & \textbf{27.71} & 57.20 & 11.40 & \textbf{23.70} & 74.30 & 9.38 & 16.88 \\
            & Blend-CL & 26.51 & \textbf{63.00} & \textbf{11.76} & 21.04 & \textbf{76.42} & \textbf{11.15} & \textbf{17.29} \\
            \midrule
            \multirow{2}{*}{LLaMA3.1-8B}
            & Blend & 28.92 & 62.00 & 9.19 & 24.74 & \textbf{76.88} & 11.88 & \textbf{17.50} \\
            & Blend-CL & \textbf{32.53} & \textbf{65.60} & \textbf{13.60} & \textbf{25.78} & 76.72 & \textbf{13.54} & 17.29 \\
            \bottomrule
\end{tabular}
}
\caption{Performance of three models on seven benchmarks with MathMixupQA based on MATH seed dataset.}
    \label{tab:math_seed_results}
\end{table*}

\begin{table*}[t!]
\centering
\resizebox{0.93\textwidth}{!}{
\begin{tabular}{c|c|cccccccccc}
\toprule
            Model & Type & AMC23 & MATH & Minerva Math & OlympiadBench & GSM8K & AIME24 & AIME25 \\
            \midrule
            \multirow{2}{*}{Qwen2.5-7B}
            & Blend & 51.81 & \textbf{78.20} & 27.21 & 39.26 & 87.49 & 35.52 & \textbf{31.98} \\
            & Blend-CL & \textbf{62.65} & 77.80 & \textbf{30.15} & \textbf{40.30} & \textbf{89.31} & \textbf{36.25} & 31.46 \\
            \midrule
            \multirow{2}{*}{InternLM2.5-7B}
            & Blend & 22.89 & \textbf{51.60} & \textbf{9.19} & 19.70 & \textbf{71.49} & \textbf{13.02} & 18.65 \\
            & Blend-CL & \textbf{27.71} & 51.40 & 8.46 & \textbf{24.00} & 70.81 & 12.50 & \textbf{20.42} \\
            \midrule
            \multirow{2}{*}{LLaMA3.1-8B}
            & Blend & 25.30 & 54.60 & \textbf{11.40} & 23.11 & \textbf{77.18} & 13.75 & 21.98 \\
            & Blend-CL & \textbf{31.33} & \textbf{56.20} & 10.66 & \textbf{26.22} & 75.13 & \textbf{14.48} & \textbf{22.19} \\
            \bottomrule
\end{tabular}
}
\caption{Performance of three models on seven benchmarks with MathMixupQA based on AMC-AIME seed dataset.}
    \label{tab:amc_aime_seed_results}
\end{table*}

\begin{table*}[t!]
\centering
\resizebox{\textwidth}{!}{
\begin{tabular}{c|ccc|cccccccc}
\toprule
Seed & Decompose & Original & Hybrid & AIME25 & AIME24 & OlympiadBench & AMC23 & MATH500 & Minerva Math & GSM8K \\
\midrule
\multirow{3}{*}{MATH}
      &       & \cmark & \cmark & 24.90 & 26.35 & 35.11 & 34.94 & 74.60 & 27.21 & 87.49 \\
      & \cmark & \cmark &       & 23.54 & 25.00 & 31.26 & 39.76 & 73.00 & 24.26 & 86.73 \\
      & \cmark &       & \cmark & 27.40 & 25.21 & 33.04 & 46.99 & 75.20 & 25.37 & 86.50 \\
\midrule
\multirow{3}{*}{ACM-AIME}
      &       & \cmark & \cmark & 24.90 & 27.19 & 33.04 & 45.78 & 71.20 & 24.26 & 87.41 \\
      & \cmark & \cmark &       & 24.17 & 23.85 & 33.78 & 40.96 & 70.80 & 24.63 & 86.96 \\
      & \cmark &       & \cmark & 27.29 & 28.13 & 34.52 & 44.58 & 72.20 & 27.21 & 85.60 \\
\bottomrule
\end{tabular}
}
\caption{Effect of data components on Qwen2.5-7B across seven benchmarks.}
\label{tab:data_components}
\end{table*}

\begin{table*}[t!]
\centering
\resizebox{\textwidth}{!}{
\begin{tabular}{c|cc|cccccccccc}
\toprule
Seed & Stage1 Data & Stage2 Data  & AIME25 & AIME24 & OlympiadBench & AMC23 & MATH500 & Minerva Math & GSM8K \\
\midrule
\multirow{3}{*}{MATH}
& Decompose & Original & 23.44 & 23.65 & 32.89 & 37.35 & 72.20 & 26.10 & 87.04 \\
& Original  & Hybrid    & 24.90 & 28.02 & 36.00 & 43.37 & 76.60 & 27.94 & 86.50 \\
& Decompose & Hybrid   & 26.35 & 26.25 & 37.78 & 44.58 & 77.00 & 28.68 & 87.49 \\
\midrule
\multirow{3}{*}{AMC-AIME}
& Decompose & Original & 24.79 & 24.69 & 35.11 & 36.14 & 71.40 & 26.10 & 89.08 \\
& Original  & Hybrid    & 26.46 & 29.58 & 36.15 & 46.99 & 75.00 & 26.10 & 86.73 \\
& Decompose & Hybrid   & 23.96 & 27.81 & 36.89 & 49.40 & 76.00 & 28.68 & 88.10 \\
\bottomrule
\end{tabular}
}
\caption{Effect of data curriculum learning on Qwen2.5-7B across seven benchmarks.}
\label{tab:data_curriculum_learning}
\end{table*}

\end{document}